%% file: neurips_2026.tex
\documentclass{article}

\PassOptionsToPackage{numbers,compress}{natbib}
\usepackage[preprint]{neurips_2026}

\usepackage[utf8]{inputenc}
\usepackage[T1]{fontenc}
\usepackage{amsmath,amssymb,amsfonts}
\usepackage{algorithm}
\usepackage{algpseudocode}
\usepackage{booktabs}
\usepackage{multirow}
\usepackage{graphicx}
\usepackage[table]{xcolor}
\usepackage{url}
\usepackage{hyperref}
\usepackage{nicefrac}
\usepackage{microtype}

\definecolor{bestbg}{HTML}{CADBB7}
\definecolor{secondbg}{HTML}{E6EEDD}

\graphicspath{{figures/}{../figures/}}
\begin{document}

\title{Latent Causal Void: Explicit Missing-Context Reconstruction for Misinformation Detection}

\author{
Hui Li\thanks{Equal contribution.} \\
School of Informatics, Xiamen University \\
Xiamen, China \\
\texttt{huilinlp@xmu.edu.cn}
\And
Zhongquan Jian\footnotemark[1] \\
School of Computer and Data Science, Minjiang University \\
Fuzhou, China
\AND
Jinsong Su\thanks{Corresponding authors.} \\
School of Informatics, Xiamen University \\
Xiamen, China
\And
Junfeng Yao\footnotemark[2] \\
School of Informatics, Xiamen University \\
Xiamen, China
}

\maketitle

\begin{abstract}
Automatic misinformation detection performs well when deception is visible in what an article explicitly states. However, some misinformation articles remain locally coherent and only become misleading once compared with contemporaneous reports that supply background facts the article omits. We study this omission-relevant setting and observe that current omission-aware approaches typically either attach retrieved context as auxiliary evidence or infer a categorical omission signal, leaving the specific missing fact implicit. We propose \emph{Latent Causal Void} (LCV), a retrieval-guided detector that explicitly reconstructs the missing fact for each target sentence and uses it as a textual cross-source relation in graph reasoning. Concretely, LCV retrieves temporally aligned context articles, asks a frozen instruction-tuned large language model to generate a short missing-context description for each sentence--article pair, and feeds the resulting relation text into a heterograph over target sentences and context articles. On the bilingual benchmark of Sheng et al., LCV improves over the strongest omission-aware baseline by $2.56$ and $2.84$ macro-F1 points on the English and Chinese splits, respectively. The results indicate that modeling the missing cross-source fact itself, rather than only attaching retrieved evidence or predicting an omission signal, is a useful representation for omission-aware misinformation detection.
\end{abstract}

\section{Introduction}

The proliferation of misinformation has severe societal consequences, motivating substantial research on automated detection~\cite{shu2017fake,zhou2020survey}. A misinformation article is a piece of news strategically crafted to deceive readers, and such deception manifests in two main forms: (i) the explicit fabrication of misleading narratives~\cite{greifeneder2020psychology}, and (ii) the implicit omission of contextual facts essential for an informed interpretation of the reported event~\cite{carson2010lying}. Existing work predominantly addresses the former, leveraging stylistic cues~\cite{xiao2024msynfd}, knowledge-augmented multimodal modeling~\cite{liu2024fkaowl,wang2024manipulated,zhu2025mfnd}, persuasion-aware reasoning~\cite{modzelewski2025pcot}, contemporaneous environment perception~\cite{sheng2022zoom}, or retrieval-based fact verification~\cite{yue2024rafts,wang2025mdpcc} to identify what is \emph{said} and falsified.

\begin{figure*}[!htbp]
    \centering
    \includegraphics[width=0.95\textwidth]{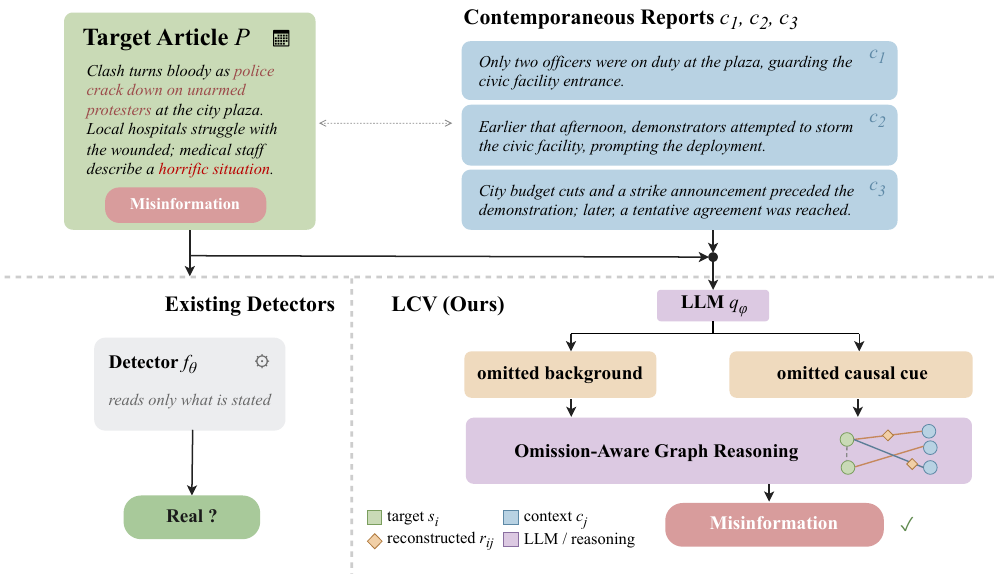}
    \caption{An omission-relevant misinformation case. The target article appears locally coherent, so a detector that reasons only over what is stated is likely to label it as real. Contemporaneous reports, however, reveal that only two officers were on duty and that the protest had attempted to storm government buildings, facts the article suppresses. LCV reconstructs these omitted facts as short cross-source relations and reasons over them in a heterograph, exposing the deception that surface-level reasoning misses.}
    \label{fig:concept}
\end{figure*}

Omission-based deception, by contrast, remains underexplored despite its damaging role. A misinformation article can be locally coherent and factually accurate at the sentence level, yet still mislead the reader by suppressing the temporal, causal, or quantitative context that would change how the event is interpreted~\cite{wang2025omigraph,sheng2022zoom}. Figure~\ref{fig:concept} illustrates this setting on a representative protest example. The target article reports street protests as a violent crackdown on unarmed demonstrators and is locally coherent enough that a detector reasoning only over its surface content is likely to accept it as real. Contemporaneous reports, however, reveal that only two officers were on duty and that the protest had attempted to storm government buildings, the very facts the article suppresses, and recovering them as cross-source relations is what allows LCV to expose the deception. Generation-side studies further show that misinformation can be obtained from real news through controlled, incremental omission rather than full fabrication~\cite{liu2025stepwise}.

Although a recent line of work begins to address omission-based deception, the modeling choice for the omission itself varies considerably. We organize existing approaches into three representational categories. \emph{(i) Evidence attachment} retrieves or attaches related documents to the target as auxiliary signals, but does not explicitly represent the difference between target and context~\cite{sheng2022zoom,yue2024rafts,shukla2025rav,xing2025evidence}. \emph{(ii) Latent omission-signal inference} predicts a categorical relation type or omission intent over target--context pairs, encoding that omission occurs without preserving its semantic content~\cite{wang2025omigraph}. \emph{(iii) Explicit missing-context reconstruction} recovers the specific fact missing from the target relative to the context and uses it as a relational feature. Only the third category directly models the missing fact itself.

To realize this third choice, we propose \emph{Latent Causal Void} (LCV), a retrieval-guided detector for omission-relevant misinformation. Given a target article, LCV retrieves temporally aligned contemporaneous reports, pairs each target sentence with each retrieved article, and uses a frozen instruction-tuned large language model (LLM) to generate a short natural-language description of the fact present in the context article but absent from the target sentence. The generated descriptions are encoded as cross-source relation embeddings and attached to a heterograph over target sentences and context articles. Relation-aware message passing then aggregates local sentence coherence, retrieved context, and document-level summary information into a single label. The key novelty of LCV is the \emph{object being modeled}: rather than predicting whether an omission relation exists or assigning it a type, LCV reconstructs the missing fact itself, allowing downstream graph reasoning to compare and weight relation semantics rather than only relation existence.

Our main contributions are summarized as follows.
\begin{itemize}
    \item \textbf{Explicit missing-context reconstruction.} We propose to model omission-based deception by directly reconstructing the missing fact as readable cross-source relation text, and we organize the design space of omission-aware methods into three representational categories of evidence attachment, latent omission-signal inference, and explicit reconstruction, in order to position this choice.
    \item \textbf{LCV framework.} We instantiate this idea as a retrieval-guided heterograph framework that combines temporally aligned context retrieval, frozen-LLM relation generation, and relation-aware graph reasoning over target sentences and context articles.
    \item \textbf{Empirical investigation.} We evaluate LCV on the public English Twitter and fact-checking benchmark and the Chinese Weibo benchmark of Sheng et al.~\cite{sheng2022zoom}, showing consistent gains of $+2.56$ and $+2.84$ macro-F1 over the strongest omission-aware baseline and clarifying the current limits of retrieval quality, relation faithfulness, and graph necessity.
\end{itemize}

\section{Related Work}

\subsection{Misinformation Detection from Presented Content}
Existing misinformation detection methods primarily focus on what is presented in the target article. Content-only methods leverage linguistic and stylistic features such as syntactic structure~\cite{xiao2024msynfd}, dual-emotion signals~\cite{zhang2021dualemo}, and persuasion-augmented reasoning~\cite{modzelewski2025pcot}. Recent advances extend this direction with knowledge-augmented large vision-language models (LVLMs)~\cite{liu2024fkaowl}, multimodal manipulation analysis~\cite{wang2024manipulated}, multimodal datasets and shallow-deep multitask learning~\cite{zhu2025mfnd}, multimodal inverse attention for fake-news detection~\cite{zhang2025mian}, causal-inference-based explanation~\cite{chen2025cife}, multi-agent cross-domain detection~\cite{li2025maro}, explainable multimodal assistants~\cite{yan2025trustvl}, contamination-aware evaluation for LLM-driven detection~\cite{xu2025ssa,xu2025triplefact}, and retrieval-augmented in-context learning over emotional cues~\cite{liu2025raemollm}.

A complementary line of work treats misinformation detection as a verification problem against external information. NEP~\cite{sheng2022zoom} compares the target against a contemporaneous news environment to capture uniqueness and popularity signals. RAFTS~\cite{yue2024rafts} prompts an LLM to generate contrastive arguments grounded in retrieved evidence. RAV~\cite{shukla2025rav} couples evidence reconnaissance with answer verification across multiple agents. MD-PCC~\cite{wang2025mdpcc} uses an external commonsense tool to detect commonsense conflicts within target content. Multi-element retrieval~\cite{chen2025birds} enriches the candidate evidence pool by aligning fine-grained elements across modalities, knowledge-graph reasoning~\cite{lourenco2026kgcraft} grounds verification in structured facts, timeline-based temporal verification~\cite{barik2025chronofact} aligns claims against their temporal context, and recent work studies evidence attribution quality~\cite{xing2025evidence}, multilingual retrieval bias~\cite{vykopal2026bias}, and security threats to automated fact-checking~\cite{schlichtkrull2025attacks}. These methods broaden the information available to the detector, but they treat retrieved context as additive evidence: the model decides whether external sources support or contradict the target, rather than encoding the specific fact missing from the target relative to that context.

\subsection{Omission-Aware Detection}
A recent line of work explicitly recognizes that misinformation can deceive through what is omitted rather than only what is fabricated, an idea long studied in the philosophy of deception~\cite{carson2010lying} and the psychology of fake news~\cite{greifeneder2020psychology}. OmiGraph~\cite{wang2025omigraph} constructs an omission-aware graph over a target article and its contextual environment, prompts an LLM to reason about omission intent for each sentence--article pair, and encodes the resulting intent as a categorical edge attribute for graph propagation. Generation-side studies show that misinformation can be derived from real news through controlled, incremental omission~\cite{liu2025stepwise}, providing further evidence that the gap between target and context is itself a useful modeling target.

These methods establish omission as an explicit modeling target, but most represent the gap through task labels, edit operations, or categorical edge attributes. In contrast, LCV reconstructs the missing fact as natural-language relation text and feeds it directly into graph reasoning, allowing the model to compare omission semantics rather than only relation existence.

\section{Method}
\begin{figure*}[!htbp]
    \centering
    \includegraphics[width=\textwidth]{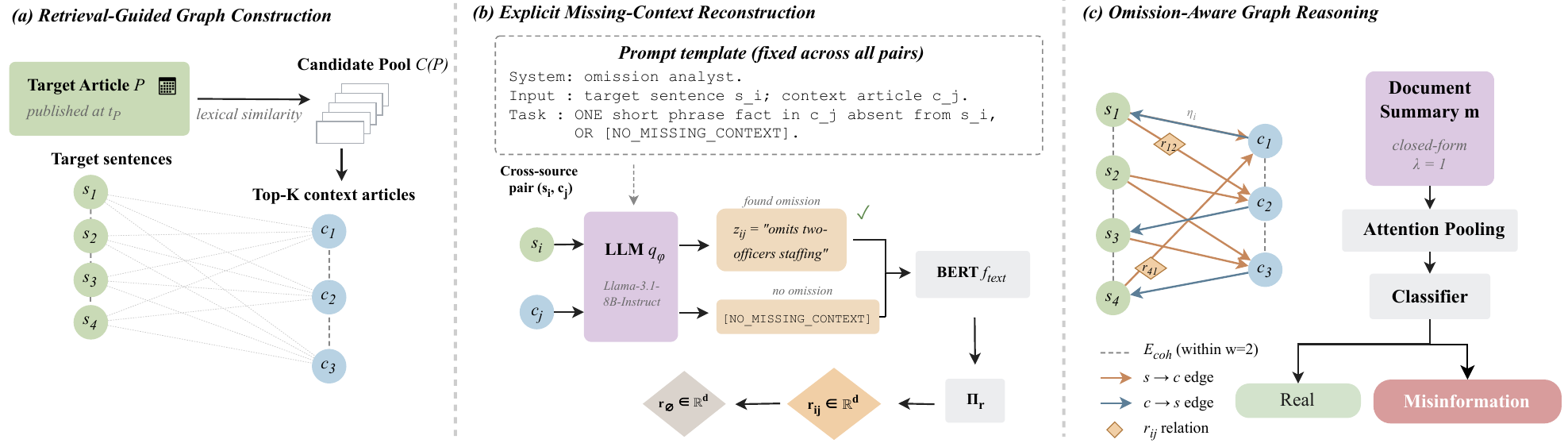}
    \caption{Overview of LCV. (a) Retrieval-guided graph construction selects the top-$K$ temporally aligned context articles from the candidate pool $\mathcal{C}(P)$ within a $\Delta$-day look-back window via TF--IDF cosine similarity, and assembles a heterograph over target sentence nodes $\mathcal{V}_s$ and context article nodes $\mathcal{V}_c$. (b) For each sentence--article pair $(s_i,c_j)\in E_{sc}$, a frozen LLM $q_\phi$ reconstructs a short missing-context description $z_{ij}$, which is encoded by the shared text encoder and projected into the cross-source relation embedding $\mathbf{r}_{ij}$. (c) Omission-aware graph reasoning propagates messages along three edge types, namely intra-target coherence edges $\mathcal{E}_{\mathrm{coh}}$, $s\!\to\!c$ and $c\!\to\!s$ cross-source edges with $\mathbf{r}_{ij}$ as edge features, and combines them with a closed-form document-level summary $\mathbf{m}$ before a binary classifier predicts $y\in\{\text{real},\text{misinfo}\}$.}
    \label{fig:framework}
\end{figure*}

\begin{figure*}[!htbp]
    \centering
    \includegraphics[width=\textwidth]{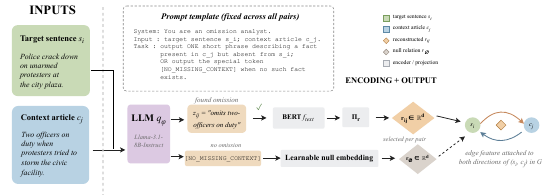}
    \caption{Construction of textual cross-source relations. For each sentence--article pair $(s_i,c_j)$, a frozen LLM $q_\phi$ is queried with a fixed prompt template that asks it to output one short phrase summarizing a fact present in $c_j$ but absent from $s_i$, or the sentinel \texttt{[NO\_MISSING\_CONTEXT]} when no such fact exists. Generated phrases $z_{ij}$ are encoded by the shared BERT backbone $f_\text{text}$ and projected by $\Pi_r$ into the cross-source relation embedding $\mathbf{r}_{ij}\in\mathbb{R}^{d}$, while \texttt{[NO\_MISSING\_CONTEXT]} outputs are mapped to a learnable null embedding $\mathbf{r}_{\varnothing}$. The resulting embeddings are attached as edge features to the heterograph used in Figure~\ref{fig:framework}(c).}
    \label{fig:edges}
\end{figure*}

We present LCV as a retrieval-guided graph learner for omission-relevant misinformation detection. Figure~\ref{fig:framework} shows the full pipeline, Figure~\ref{fig:edges} zooms in on the relation-construction stage, and Figure~\ref{fig:reason} details the graph reasoning module. The method consists of five components: problem formulation (Sec.~\ref{sec:formulation}), retrieval-guided graph construction (Sec.~\ref{sec:graph}), explicit missing-context reconstruction (Sec.~\ref{sec:recon}), omission-aware graph reasoning (Sec.~\ref{sec:reason}), and prediction with optimization (Sec.~\ref{sec:predict}).

\subsection{Problem Formulation}
\label{sec:formulation}
Let $P$ denote a target candidate article published at time $t_P$, with binary label $y\in\{0,1\}$, where $y\!=\!1$ denotes misinformation. The article is segmented into a sentence sequence $S(P)\!=\!\{s_1,\ldots,s_n\}$ with $n$ bounded by a sentence budget. The candidate-context pool $\mathcal{C}(P)$ collects all articles published in a fixed look-back window $[t_P-\Delta,t_P)$, where $\Delta$ is the temporal radius. From $\mathcal{C}(P)$, a retrieval module selects the top-$K$ most relevant articles $\operatorname{Ret}_K(P)\!=\!\{c_1,\ldots,c_K\}$, where $K$ is the retrieval budget. The cross-source pair set is $E_{sc}\!=\!\{(s_i,c_j)\mid s_i\in S(P),\,c_j\in\operatorname{Ret}_K(P)\}$. The prediction problem is to learn a classifier $f_\theta:(P,S(P),\operatorname{Ret}_K(P),E_{sc})\!\mapsto\!y$, where the central modeling challenge is to recover cross-source facts that are present in $\operatorname{Ret}_K(P)$ but absent from $S(P)$, rather than merely aggregating observed text. The hyperparameters $\Delta$, $n$, and $K$ are fixed at training and reported in Section~\ref{sec:setup}.

\subsection{Retrieval-Guided Graph Construction}
\label{sec:graph}
Following Sheng et al.~\cite{sheng2022zoom}, LCV uses TF--IDF cosine similarity between $P$ and each candidate $c\in\mathcal{C}(P)$ to rank candidates and select the top-$K$ as $\operatorname{Ret}_K(P)$. Target sentences and retrieved articles are then assembled into a heterogeneous graph $\mathcal{G}=(\mathcal{V}_s,\mathcal{V}_c,\mathcal{E}_{\mathrm{coh}},E_{sc},\mathbf{R})$, where $\mathcal{V}_s\!=\!\{s_i\}_{i=1}^n$ is the set of target-sentence nodes, $\mathcal{V}_c\!=\!\{c_j\}_{j=1}^K$ is the set of context-article nodes, $\mathcal{E}_{\mathrm{coh}}$ is the set of intra-target sentence-coherence edges defined in Sec.~\ref{sec:reason}, $E_{sc}$ is the set of cross-source sentence--article edges, and $\mathbf{R}\!=\!\{\mathbf{r}_{ij}\}$ collects the cross-source relation embeddings reconstructed in Sec.~\ref{sec:recon}.

Let $f_\text{text}(\cdot)$ denote the shared frozen BERT~\cite{devlin2019bert} text encoder used for sentences, articles, relation text, and full documents, with output dimension $d_0\!=\!768$. Independent single-layer linear projections $\Pi_s,\Pi_c,\Pi_g,\Pi_r:\mathbb{R}^{d_0}\!\rightarrow\!\mathbb{R}^{d}$ map this common text space into the graph node, document, and relation spaces with hidden size $d$. The initial node features are
\begin{equation}
\mathbf{h}_i^{(0)}\!=\!\Pi_s f_\text{text}(s_i),\quad
\mathbf{u}_j^{(0)}\!=\!\Pi_c f_\text{text}(c_j),\quad
\mathbf{g}^{(0)}\!=\!\Pi_g f_\text{text}(P),
\end{equation}
where $\mathbf{h}_i^{(0)},\mathbf{u}_j^{(0)},\mathbf{g}^{(0)}\!\in\!\mathbb{R}^{d}$ are the initial sentence, context, and document representations.

\subsection{Explicit Missing-Context Reconstruction}
\label{sec:recon}
The core stage of LCV reconstructs, for each cross-source pair $(s_i,c_j)\!\in\!E_{sc}$, a short natural-language description of the fact present in $c_j$ but absent from $s_i$. We elicit this description from a frozen instruction-tuned LLM $q_\phi$:
\begin{equation}
z_{ij}=q_\phi(s_i,c_j),
\end{equation}
where $\phi$ remains fixed throughout training. The prompt asks $q_\phi$ to output either one short phrase summarizing a fact present in $c_j$ but absent from $s_i$, or the special token \texttt{[NO\_MISSING\_CONTEXT]} when no such fact exists. The exact prompt template, decoding configuration, and caching policy are described in Section~\ref{sec:setup}.

The reconstructed text $z_{ij}$ is encoded into the cross-source relation embedding $\mathbf{r}_{ij}\!=\!\Pi_r f_\text{text}(z_{ij})\!\in\!\mathbb{R}^{d}$ and attached to both directions of the corresponding sentence--article edge in $\mathcal{G}$. Pairs whose output is \texttt{[NO\_MISSING\_CONTEXT]} are mapped to a learnable null relation embedding $\mathbf{r}_{\varnothing}\!\in\!\mathbb{R}^{d}$, which preserves edge connectivity while signaling the absence of an informative omission. This design lets retrieved context contribute to graph reasoning not merely by being attached to the target article, but through an explicit, readable representation of the fact missing from the target relative to that context.

\subsection{Omission-Aware Graph Reasoning}
\label{sec:reason}
\begin{figure*}[!htbp]
    \centering
    \includegraphics[width=\textwidth]{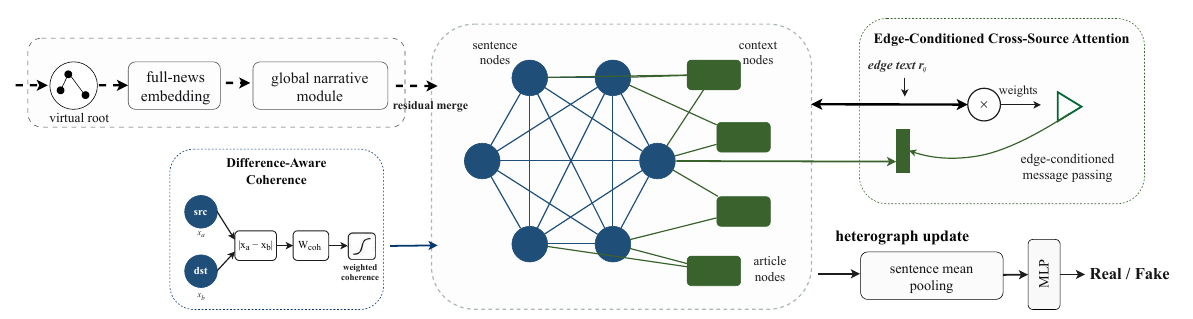}
    \caption{Graph-centric view of omission-aware reasoning. The \emph{Difference-Aware Coherence} block (left) turns each coherence edge $(s_i,s_k)\!\in\!\mathcal{E}_{\mathrm{coh}}$ into the weight $\kappa_{\mathrm{coh}}(\mathbf{x}_a^{(\ell)},\mathbf{x}_b^{(\ell)})\!=\!\exp(-\lVert\mathbf{W}_{\mathrm{coh}}(\mathbf{x}_a^{(\ell)}-\mathbf{x}_b^{(\ell)})\rVert_2^2)$ of Eq.~\eqref{eq:diffweight}. The \emph{Edge-Conditioned Cross-Source Attention} block (right) fuses node states with the relation embedding $\mathbf{r}_{ij}$ via the bilinear score of Eq.~\eqref{eq:psi}, yielding edge-conditioned messages. Heterograph update (center) alternates the two message types over $L$ layers; a virtual root produces the full-news embedding $\mathbf{g}^{(0)}$, which is residually merged with the propagated states into the global-narrative summary $\mathbf{m}$ of Eq.~\eqref{eq:global}. Sentence mean pooling of the refined states $\{\tilde{\mathbf{h}}_i\}$ feeds the MLP classifier that predicts \texttt{Real}/\texttt{Fake}.}
    \label{fig:reason}
\end{figure*}

The graph reasoner operates on $\mathcal{G}$ with a unified node hidden state $\mathbf{x}_a^{(\ell)}\!\in\!\mathbb{R}^{d}$ for every node $a\!\in\!\mathcal{V}_s\!\cup\!\mathcal{V}_c$ at layer $\ell$, where $\mathbf{x}_a^{(0)}\!=\!\mathbf{h}_i^{(0)}$ when $a\!=\!s_i$ and $\mathbf{x}_a^{(0)}\!=\!\mathbf{u}_j^{(0)}$ when $a\!=\!c_j$. The relation-type indicator $\rho(a,b)\!\in\!\{\text{coh},s\!\to\!c,c\!\to\!s\}$ identifies which of three edge types connects $a$ and $b$. Figure~\ref{fig:reason} illustrates the three reasoning components described below and their interaction with the document-level summary.

\paragraph{Local sentence coherence.}
Target sentences are connected through a local window graph. With window size $w$, the coherence edge set is $\mathcal{E}_{\mathrm{coh}}\!=\!\{(s_i,s_k)\mid 0\!<\!|i-k|\!\le\!w\}$. For each coherence edge, LCV computes a difference-aware compatibility score
\begin{equation}
\kappa_{\mathrm{coh}}(\mathbf{x}_a^{(\ell)},\mathbf{x}_b^{(\ell)})
=\exp\!\bigl(-\bigl\lVert \mathbf{W}_{\mathrm{coh}}(\mathbf{x}_a^{(\ell)}-\mathbf{x}_b^{(\ell)})\bigr\rVert_2^2\bigr),
\label{eq:diffweight}
\end{equation}
where $\mathbf{W}_{\mathrm{coh}}\!\in\!\mathbb{R}^{d\times d}$ is a learnable projection. Neighboring sentences receive a lower coherence score when their embeddings differ sharply, helping the model surface local shifts often associated with omission-based framing. The coherence score enters the unified attention formulation below as the unnormalized score $\psi_{ab}^{(\ell)}\!=\!\log\kappa_{\mathrm{coh}}(\mathbf{x}_a^{(\ell)},\mathbf{x}_b^{(\ell)})$ for edges of type $\rho(a,b)\!=\!\text{coh}$.

\paragraph{Relation-aware cross-source attention.}
For cross-source edges, the unnormalized attention score is computed jointly from node states and the textual relation embedding:
\begin{equation}
\psi_{ab}^{(\ell)}
=\langle \mathbf{x}_a^{(\ell)},\mathbf{Q}_{\rho(a,b)}^{(\ell)}\mathbf{x}_b^{(\ell)}\rangle
+\langle \mathbf{r}_{ab},\mathbf{T}_{\rho(a,b)}^{(\ell)}\mathbf{r}_{ab}\rangle,
\label{eq:psi}
\end{equation}
where $\mathbf{Q}_{\rho}^{(\ell)},\mathbf{T}_{\rho}^{(\ell)}\!\in\!\mathbb{R}^{d\times d}$ are learnable relation-specific bilinear forms; $\mathbf{Q}_{\rho}^{(\ell)}$ scores node--node compatibility and $\mathbf{T}_{\rho}^{(\ell)}$ scores how strongly the relation text $\mathbf{r}_{ab}$ supports the edge. The attention weight and node update are
\begin{equation}
\pi_{ab}^{(\ell)}
=\frac{\exp(\psi_{ab}^{(\ell)})}{\sum_{b'\in\mathcal{N}(a)}\exp(\psi_{ab'}^{(\ell)})},\qquad
\mathbf{x}_a^{(\ell+1)}
=\sigma\!\left(\sum_{b\in\mathcal{N}(a)}\pi_{ab}^{(\ell)}\,\mathbf{W}_{\rho(a,b)}^{(\ell)}\mathbf{x}_b^{(\ell)}\right),
\label{eq:cfattn}
\end{equation}
where $\mathcal{N}(a)$ denotes the neighbors of $a$ in $\mathcal{G}$, $\mathbf{W}_{\rho}^{(\ell)}\!\in\!\mathbb{R}^{d\times d}$ is the relation-specific message transform, and $\sigma(\cdot)$ is the ReLU non-linearity. After $L$ propagation layers, the sentence states $\mathbf{h}_i^{(L)}\!\equiv\!\mathbf{x}_{s_i}^{(L)}$ carry both target-internal coherence and relation-aware cross-source evidence.

\paragraph{Document-aware global summary.}
LCV maintains a closed-form document-level summary $\mathbf{m}=(\sum_{i=1}^n \mathbf{h}_i^{(L)}+\lambda\mathbf{g}^{(0)})/(n+\lambda)$, where $\lambda\!\ge\!0$ is a scalar mixing coefficient that balances the propagated sentence states and the original document embedding. Each sentence representation is then refined via a learned attention vector $\mathbf{a}\!\in\!\mathbb{R}^{d}$:
\begin{equation}
\tilde{\mathbf{h}}_i=\mathbf{h}_i^{(L)}+\eta_i\mathbf{m},\qquad
\eta_i=\frac{\exp(\langle \mathbf{a},\mathbf{h}_i^{(L)}\rangle)}{\sum_{k=1}^n\exp(\langle \mathbf{a},\mathbf{h}_k^{(L)}\rangle)}.
\label{eq:global}
\end{equation}
This injection contextualizes segment-level omission cues within the overall narrative, while avoiding the over-smoothing risk of stacking additional message-passing layers.

\begin{algorithm}[!htbp]
	\caption{Training LCV.}
	\label{alg:lcv}
	\begin{algorithmic}[1]
		\Require Article $(P,y)$, retrieval budget $K$, propagation depth $L$, mixing coefficient $\lambda$
		\State Retrieve $\operatorname{Ret}_K(P)$ from $\mathcal{C}(P)$ via TF--IDF cosine similarity
		\State Segment $P$ into sentences $S(P)$ and form the cross-source pair set $E_{sc}$
		\State Generate $z_{ij}\!=\!q_\phi(s_i,c_j)$ and encode nodes/relations $\{\mathbf{h}_i^{(0)},\mathbf{u}_j^{(0)},\mathbf{g}^{(0)},\mathbf{r}_{ij}\}$
		\For{$\ell=0,\ldots,L-1$}
		\State Update all $\mathbf{x}_a^{(\ell+1)}$ via the omission-aware graph operator
		\EndFor
		\State Compute the document summary $\mathbf{m}$, refine $\{\tilde{\mathbf{h}}_i\}$, pool to $\mathbf{p}$, and predict $\hat{y}$
		\State Update parameters by descending $\nabla_\theta \mathcal{L}_\text{cls}$
	\end{algorithmic}
\end{algorithm}

\subsection{Prediction and Optimization}
\label{sec:predict}
The target-side graph is summarized by attention-pooling the refined sentence representations: $\mathbf{p}=\sum_{i=1}^n \beta_i\tilde{\mathbf{h}}_i$ with $\beta_i\!=\!\mathrm{softmax}_i(\langle \mathbf{b},\tilde{\mathbf{h}}_i\rangle)$ over $i$, where $\mathbf{b}\!\in\!\mathbb{R}^{d}$ is a learnable pooling vector. The label distribution is $\Pr(y\mid P,\operatorname{Ret}_K(P))\!=\!\operatorname{softmax}(\mathbf{W}_o\mathbf{p}+\mathbf{b}_o)$, and the model is trained end-to-end with the cross-entropy loss $\mathcal{L}_\text{cls}=\mathbb{E}_{(P,y)\sim\mathcal{D}}[-\log\Pr(y\mid P,\operatorname{Ret}_K(P))]$.

\section{Experiments}
\label{sec:exp}

\subsection{Datasets and Evaluation Protocol}
Following Sheng et al.~\cite{sheng2022zoom}, we use one English dataset collected from Twitter and fact-checking websites, and one Chinese dataset collected from Weibo. Both are coupled with contemporaneous mainstream-news corpora as the candidate-context pool. Table~\ref{tab:dataset-overview} summarizes the two datasets, and Table~\ref{tab:data} reports processed split statistics for the English split. We report macro-F1, accuracy, and per-class F1, following prior omission-aware comparisons.

\begin{table}[!htbp]
    \caption{Datasets used in our experiments.}
    \label{tab:dataset-overview}
    \centering
    \small
    \begin{tabular}{p{0.20\linewidth}p{0.24\linewidth}p{0.42\linewidth}}
        \toprule
        Dataset & Target source & Context source \\
        \midrule
        English~\cite{sheng2022zoom} & Twitter and fact-checking websites & Contemporaneous mainstream-news articles within a seven-day window \\
        Chinese~\cite{sheng2022zoom} & Weibo & Contemporaneous mainstream-news articles within a seven-day window \\
        \bottomrule
    \end{tabular}
\end{table}

\begin{table}[!htbp]
    \caption{Processed split statistics for the English dataset used in our experiments.}
    \label{tab:data}
    \centering
    \resizebox{0.98\linewidth}{!}{%
    \begin{tabular}{lrrrrrr}
        \toprule
        Split & \#Articles & Real & Misinfo. & Avg. \#sentences & Avg. \#context articles & Avg. \#cross edges \\
        \midrule
        Train & 3900 & 1976 & 1924 & 1.378 & 3.000 & 4.134 \\
        Val   & 1294 & 656  & 638  & 1.385 & 3.000 & 4.155 \\
        Test  & 1289 & 661  & 628  & 1.463 & 3.000 & 4.389 \\
        \bottomrule
    \end{tabular}}
\end{table}

\subsection{Implementation Details}
\label{sec:setup}
\paragraph{Retrieval and graph.}
LCV retrieves the top-$K\!=\!3$ context articles within a $\Delta\!=\!7$-day look-back window using TF--IDF cosine similarity. Target articles are segmented into at most $n\!=\!10$ sentences. The propagation depth is $L\!=\!2$, the coherence window is $w\!=\!2$, and the global mixing coefficient is $\lambda\!=\!1$.

\paragraph{Encoders.}
Sentences, articles, recovered missing-context phrases, and full documents are encoded with a shared frozen BERT backbone~\cite{devlin2019bert}: \texttt{bert-base-uncased} for English and \texttt{bert-base-chinese} for Chinese. The text encoder output dimension is $d_0\!=\!768$ and the graph hidden size is $d\!=\!256$.

\paragraph{Missing-context LLM.}
Missing-context phrases are generated offline with \texttt{Llama-3.1-8B-Instruct}~\cite{dubey2024llama} using greedy decoding with temperature $0$ and a budget of $16$ tokens. The prompt asks the LLM to output either one short phrase summarizing a fact present in the context article but absent from the target sentence, or the special token \texttt{[NO\_MISSING\_CONTEXT]} when no such fact exists; sentence and article inputs are truncated to $32$ and $256$ tokens. Outputs are post-processed by keeping only the first generated line and normalizing empty outputs or token variants to \texttt{[NO\_MISSING\_CONTEXT]}. Generation is performed once per pair and cached.

\paragraph{Training.}
We train with Adam at learning rate $2\!\times\!10^{-5}$, batch size $16$, dropout $0.1$, at most $30$ epochs, early-stopping patience $7$, and three fixed random seeds. Checkpoints are selected by validation macro-F1.

\subsection{Baselines}
We compare LCV against three groups of baselines. \emph{Content-only methods}: BERT~\cite{devlin2019bert}, DualEmo~\cite{zhang2021dualemo}, MSynFD~\cite{xiao2024msynfd}, an LLM-only classifier using the same generator backbone~\cite{dubey2024llama}, and PCoT~\cite{modzelewski2025pcot}. \emph{External information-aware methods}: NEP~\cite{sheng2022zoom}, MD-PCC~\cite{wang2025mdpcc}, RAV~\cite{shukla2025rav}, and RAFTS~\cite{yue2024rafts}. \emph{Omission-aware graph methods}: OmiGraph~\cite{wang2025omigraph}. All baselines except OmiGraph are re-run in our environment with three fixed seeds and the same data splits; OmiGraph numbers are taken from the strongest published configuration~\cite{wang2025omigraph}, since a faithful local re-implementation was unavailable.

\subsection{Main Results}
Table~\ref{tab:compare} reports the main comparison. LCV achieves the best macro-F1, accuracy, and per-class F1 on both datasets, improving over OmiGraph, the strongest omission-aware baseline, by $+2.56$ macro-F1 on the English dataset and $+2.84$ macro-F1 on the Chinese dataset. Paired $t$-tests over three runs show that the improvement of LCV over OmiGraph is significant at the $0.005$ level on every metric on both datasets. The gain over external information-aware methods is even larger, supporting our central claim that retrieved context becomes most useful when the model explicitly reconstructs the fact missing from the target relative to that context, rather than treating retrieval as additive evidence.

\begin{table*}[!htbp]
    \caption{Main comparison on the English and Chinese datasets. Higher is better. Re-run baselines and LCV are reported as mean$\pm$standard deviation over three runs. Entries marked with~$\dagger$ are taken from the published paper and therefore do not include re-run variance. $\ast$ indicates that the improvement of LCV over OmiGraph is statistically significant at the $0.005$ level under a paired $t$-test.}
    \label{tab:compare}
    \centering
    \footnotesize
    \resizebox{\textwidth}{!}{%
    \begin{tabular}{llcccccccc}
        \toprule
        & & \multicolumn{4}{c}{English} & \multicolumn{4}{c}{Chinese} \\
        \cmidrule(lr){3-6} \cmidrule(lr){7-10}
        Category & Method & macF1 & Acc & F1-real & F1-misinfo & macF1 & Acc & F1-real & F1-misinfo \\
        \midrule
        \multirow{5}{*}{Content-only}
        & BERT~\cite{devlin2019bert} & 0.7111$\pm$.0032 & 0.7135$\pm$.0021 & 0.7367$\pm$.0035 & 0.7025$\pm$.0097 & 0.7851$\pm$.0014 & 0.7921$\pm$.0016 & 0.8240$\pm$.0018 & 0.7461$\pm$.0012 \\
        & DualEmo~\cite{zhang2021dualemo} & 0.7194$\pm$.0024 & 0.7200$\pm$.0021 & 0.7322$\pm$.0013 & 0.7065$\pm$.0048 & 0.7958$\pm$.0033 & 0.8003$\pm$.0029 & 0.8262$\pm$.0024 & 0.7655$\pm$.0056 \\
        & MSynFD~\cite{xiao2024msynfd} & 0.7317$\pm$.0018 & 0.7319$\pm$.0017 & 0.7324$\pm$.0103 & 0.7309$\pm$.0083 & 0.8054$\pm$.0052 & 0.8089$\pm$.0048 & 0.8315$\pm$.0034 & 0.7793$\pm$.0074 \\
        & LLM~\cite{dubey2024llama} & 0.5556$\pm$.0002 & 0.5779$\pm$.0000 & 0.4561$\pm$.0002 & 0.6552$\pm$.0013 & 0.6992$\pm$.0001 & 0.7110$\pm$.0001 & 0.6397$\pm$.0001 & 0.7588$\pm$.0001 \\
        & PCoT~\cite{modzelewski2025pcot} & 0.6508$\pm$.0011 & 0.6509$\pm$.0001 & 0.6434$\pm$.0000 & 0.6481$\pm$.0003 & 0.8020$\pm$.0001 & 0.8041$\pm$.0002 & 0.7812$\pm$.0001 & 0.8227$\pm$.0001 \\
        \midrule
        \multirow{4}{*}{External info.-aware}
        & NEP~\cite{sheng2022zoom} & 0.7274$\pm$.0004 & 0.7278$\pm$.0005 & 0.7383$\pm$.0011 & 0.7165$\pm$.0005 & 0.8288$\pm$.0010 & 0.8311$\pm$.0012 & 0.8486$\pm$.0012 & 0.8090$\pm$.0017 \\
        & MD-PCC~\cite{wang2025mdpcc} & 0.7227$\pm$.0028 & 0.7243$\pm$.0031 & 0.7434$\pm$.0048 & 0.7021$\pm$.0017 & 0.8168$\pm$.0022 & 0.8205$\pm$.0026 & 0.8427$\pm$.0033 & 0.7909$\pm$.0012 \\
        & RAV~\cite{shukla2025rav} & 0.7189$\pm$.0020 & 0.7197$\pm$.0019 & 0.7336$\pm$.0037 & 0.7041$\pm$.0050 & 0.7930$\pm$.0018 & 0.7980$\pm$.0011 & 0.8252$\pm$.0016 & 0.7608$\pm$.0047 \\
        & RAFTS~\cite{yue2024rafts} & 0.6016$\pm$.0005 & 0.6049$\pm$.0006 & 0.6019$\pm$.0010 & 0.6208$\pm$.0009 & 0.7427$\pm$.0003 & 0.7580$\pm$.0002 & 0.8055$\pm$.0007 & 0.6800$\pm$.0004 \\
        \midrule
        \multirow{2}{*}{Omission-aware}
        & OmiGraph~\cite{wang2025omigraph}$^{\dagger}$ & \cellcolor{secondbg}0.7608 & \cellcolor{secondbg}0.7647 & \cellcolor{secondbg}0.7586 & \cellcolor{secondbg}0.7528 & \cellcolor{secondbg}0.8585 & \cellcolor{secondbg}0.8596 & \cellcolor{secondbg}0.8711 & \cellcolor{secondbg}0.8460 \\
        & \textbf{LCV (ours)} & \cellcolor{bestbg}\textbf{0.7864$\pm$.0023}$^{\ast}$ & \cellcolor{bestbg}\textbf{0.8084$\pm$.0026}$^{\ast}$ & \cellcolor{bestbg}\textbf{0.7926$\pm$.0029}$^{\ast}$ & \cellcolor{bestbg}\textbf{0.7802$\pm$.0032}$^{\ast}$ & \cellcolor{bestbg}\textbf{0.8869$\pm$.0021}$^{\ast}$ & \cellcolor{bestbg}\textbf{0.8832$\pm$.0019}$^{\ast}$ & \cellcolor{bestbg}\textbf{0.9011$\pm$.0024}$^{\ast}$ & \cellcolor{bestbg}\textbf{0.8732$\pm$.0026}$^{\ast}$ \\
        \bottomrule
    \end{tabular}}
\end{table*}

\subsection{Ablation Study}
We isolate the main design choices in LCV with three ablations against the full model: (i) \textit{w/o Retrieved Context} disables context retrieval, leaving only target-internal sentence reasoning; (ii) \textit{Structural Edges Only} keeps the cross-source connectivity but replaces the textual relation embeddings $\mathbf{r}_{ij}$ with uniform learnable structural edge attributes; (iii) \textit{w/o Global Summary} removes the document-level summary $\mathbf{m}$ and uses only $\mathbf{h}_i^{(L)}$ for prediction. Table~\ref{tab:ablation} shows that the full model performs best on both datasets. Removing retrieved context drops macro-F1 by $\sim4.4$ points (English) and $\sim4.3$ points (Chinese); replacing textual relations with structural edges drops macro-F1 by $\sim3.0$ and $\sim2.7$ points; and removing the global summary drops macro-F1 by $\sim1.4$ and $\sim1.2$ points. Omission-driven detection thus depends most strongly on retrieved context, benefits substantially from semantic omission relations beyond bare connectivity, and still gains from document-level aggregation.

\begin{table}[!htbp]
    \caption{Ablation study on the English and Chinese datasets. Higher is better.}
    \label{tab:ablation}
    \centering
    \small
    \begin{tabular}{lcccc}
        \toprule
        & \multicolumn{2}{c}{English} & \multicolumn{2}{c}{Chinese} \\
        \cmidrule(lr){2-3} \cmidrule(lr){4-5}
        Variant & Macro-F1 & Acc. & Macro-F1 & Acc. \\
        \midrule
        Full LCV & \cellcolor{bestbg}\textbf{0.7864} & \cellcolor{bestbg}\textbf{0.8084} & \cellcolor{bestbg}\textbf{0.8869} & \cellcolor{bestbg}\textbf{0.8832} \\
        \quad \textit{w/o} Retrieved Context & 0.7421 & 0.7618 & 0.8443 & 0.8431 \\
        \quad Structural Edges Only & 0.7568 & 0.7789 & 0.8597 & 0.8574 \\
        \quad \textit{w/o} Global Summary & \cellcolor{secondbg}0.7729 & \cellcolor{secondbg}0.7936 & \cellcolor{secondbg}0.8746 & \cellcolor{secondbg}0.8718 \\
        \bottomrule
    \end{tabular}
\end{table}

\subsection{Case Study}
\begin{figure*}[!htbp]
    \centering
    \includegraphics[width=\textwidth]{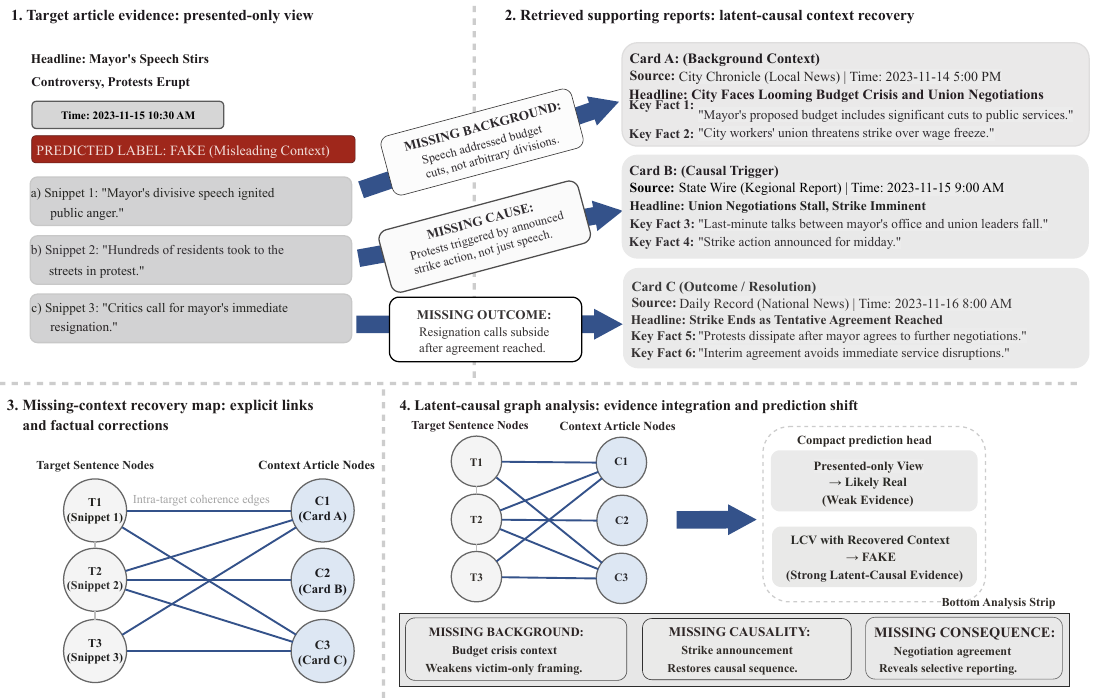}
    \caption{Qualitative LCV prediction on a selective-news-reporting case from the English dataset. Panel~1 contains the target article with three sentences $\{s_i\}$ that are locally coherent under a presented-only view. Panel~2 lists the retrieved context articles $\{c_j\}$ that supply the omitted background, causal trigger, and outcome. Panel~3 visualizes the reconstructed missing-context relations $z_{ij}$ as dark sentence--article edges in the heterograph $\mathcal{G}$, with light edges denoting intra-target coherence in $\mathcal{E}_{\mathrm{coh}}$. Panel~4 shows the same bipartite structure consumed by the omission-aware reasoner of Section~\ref{sec:reason}: the compact prediction head moves from a weak \texttt{Real} decision under the presented-only view to a confident \texttt{Fake} decision once the recovered relations are integrated, with the bottom strip attributing the shift to missing background, missing causality, and missing consequence.}
    \label{fig:case-study}
\end{figure*}

Figure~\ref{fig:case-study} traces a representative selective-reporting case from the English dataset through the four stages of LCV. Panel~1 shows the \emph{presented-only view}: a headline that frames a mayoral speech as the cause of street protests and calls for resignation. Read in isolation, the three target sentences $\{s_i\}$ are locally coherent and contain no internal contradiction, so a content-only detector lacks any surface signal that the article is misleading. Panel~2 shows three retrieved context articles $\{c_j\}$, ordered by retrieval, that together cover the budget-crisis background, the stalled union negotiations that triggered the protests, and the later tentative agreement. None of these articles individually contradicts the target; their relevance lies in what they make available that the target omits.

Panel~3 visualizes the reconstructed missing-context relations as labeled sentence--article edges $(s_i,c_j,z_{ij})$ produced by $q_\phi$ in Section~\ref{sec:recon}. Each dark edge in the bipartite map corresponds to a non-null relation phrase such as a missing budget-cut background, a missing strike-announcement cause, or a missing later-agreement outcome, while light edges denote intra-target coherence in $\mathcal{E}_{\mathrm{coh}}$. The structure makes explicit what is missing rather than merely that the target and a context article are related, which distinguishes LCV from approaches that attach evidence without reconstructing its semantic content. Panel~4 shows the prediction shift induced by these textual edges. Without recovered context, the compact prediction head assigns the article a weak \texttt{Real} score consistent with its local coherence; conditioning on the relation-augmented graph reverses the decision to \texttt{Fake} with strong latent-causal evidence. The bottom strip summarizes the three categories of omission, namely missing background, missing causality, and missing consequence, that drive this shift, illustrating how explicit textual cross-source edges turn locally plausible but selectively reported content into a detectable omission pattern.

% \subsection{Discussion and Limitations}
% LCV depends on upstream retrieval and offline relation generation. TF--IDF retrieval can miss semantically relevant but lexically distant evidence, and pairwise relation generation dominates preprocessing cost. The current experiments validate the downstream usefulness, but not the full factual faithfulness, of generated relations, and they do not yet isolate graph benefit against strong non-graph consumers of the same generated text. Future work should audit relation quality directly, study confidence-aware edge weighting, test robustness under noisy or adversarial retrieval, and measure end-to-end system cost under larger retrieval pools.

\section{Conclusion}
This paper presents LCV, a retrieval-guided misinformation detector that models omission by explicitly reconstructing missing context as textual cross-source relations. By combining temporally aligned context retrieval, missing-context reconstruction, and heterogeneous graph reasoning, LCV outperforms strong content-only, external information-aware, and omission-aware baselines on both English and Chinese benchmarks. The results suggest that misinformation detection should account not only for what an article states explicitly, but also for what relevant context it omits.

\newpage
\input{checklist.tex}

\end{document}

%% file: checklist.tex
\section*{NeurIPS Paper Checklist}

\begin{enumerate}

\item {\bf Claims}
    \item[] Question: Do the main claims made in the abstract and introduction accurately reflect the paper's contributions and scope?
    \item[] Answer: \answerYes{}
    \item[] Justification: The abstract and Introduction summarize the representational claim, the method, and the empirical scope, and the experimental sections do not claim more than the reported benchmark evidence supports.

\item {\bf Limitations}
    \item[] Question: Does the paper discuss the limitations of the work performed by the authors?
    \item[] Answer: \answerYes{}
    \item[] Justification: Section ``Discussion and Limitations'' explicitly discusses retrieval bottlenecks, lightweight graph structure, and robustness limits.

\item {\bf Theory assumptions and proofs}
    \item[] Question: For each theoretical result, does the paper provide the full set of assumptions and a complete (and correct) proof?
    \item[] Answer: \answerNA{}
    \item[] Justification: The paper does not present formal theorems or proofs; it is an empirical modeling paper.

    \item {\bf Experimental result reproducibility}
    \item[] Question: Does the paper fully disclose all the information needed to reproduce the main experimental results of the paper to the extent that it affects the main claims and/or conclusions of the paper (regardless of whether the code and data are provided or not)?
    \item[] Answer: \answerNo{}
    \item[] Justification: The paper reports key settings in Method and Experimental Setup, but it does not provide the full preprocessing scripts, exact prompt text, or a complete reproducibility package in the submission.

\item {\bf Open access to data and code}
    \item[] Question: Does the paper provide open access to the data and code, with sufficient instructions to faithfully reproduce the main experimental results, as described in supplemental material?
    \item[] Answer: \answerNo{}
    \item[] Justification: The submission does not provide anonymized code or a release package; it only describes datasets, settings, and citations to the assets used.

\item {\bf Experimental setting/details}
    \item[] Question: Does the paper specify all the training and test details (e.g., data splits, hyperparameters, how they were chosen, type of optimizer) necessary to understand the results?
    \item[] Answer: \answerYes{}
    \item[] Justification: The Method and Experimental Setup sections specify the datasets, retrieval window, model components, major hyperparameters, and evaluation metrics used in the reported results.

\item {\bf Experiment statistical significance}
    \item[] Question: Does the paper report error bars suitably and correctly defined or other appropriate information about the statistical significance of the experiments?
    \item[] Answer: \answerYes{}
    \item[] Justification: The main comparison reports re-run baselines and LCV as mean$\pm$standard deviation over three fixed seeds, and the improvement of LCV over the strongest omission-aware baseline is verified by a paired $t$-test, with significance at the $0.005$ level marked in the table.

\item {\bf Experiments compute resources}
    \item[] Question: For each experiment, does the paper provide sufficient information on the computer resources (type of compute workers, memory, time of execution) needed to reproduce the experiments?
    \item[] Answer: \answerNo{}
    \item[] Justification: The paper discusses preprocessing cost qualitatively in the limitations section, but it does not report hardware, memory, or wall-clock runtime details.
    
\item {\bf Code of ethics}
    \item[] Question: Does the research conducted in the paper conform, in every respect, with the NeurIPS Code of Ethics \url{https://neurips.cc/public/EthicsGuidelines}?
    \item[] Answer: \answerYes{}
    \item[] Justification: The study uses public benchmark data and standard evaluation protocols, does not involve human subjects experiments, and does not describe any deviation from the NeurIPS Code of Ethics.

\item {\bf Broader impacts}
    \item[] Question: Does the paper discuss both potential positive societal impacts and negative societal impacts of the work performed?
    \item[] Answer: \answerNo{}
    \item[] Justification: The current submission does not include a dedicated broader-impact discussion, although the task domain is misinformation detection and clearly has societal relevance.
    
\item {\bf Safeguards}
    \item[] Question: Does the paper describe safeguards that have been put in place for responsible release of data or models that have a high risk for misuse (e.g., pre-trained language models, image generators, or scraped datasets)?
    \item[] Answer: \answerNA{}
    \item[] Justification: The paper does not release a new high-risk model or dataset as part of the submission.

\item {\bf Licenses for existing assets}
    \item[] Question: Are the creators or original owners of assets (e.g., code, data, models), used in the paper, properly credited and are the license and terms of use explicitly mentioned and properly respected?
    \item[] Answer: \answerNo{}
    \item[] Justification: The paper cites the main datasets, models, and baselines, but it does not explicitly document licenses or terms of use for all existing assets.

\item {\bf New assets}
    \item[] Question: Are new assets introduced in the paper well documented and is the documentation provided alongside the assets?
    \item[] Answer: \answerNA{}
    \item[] Justification: The submission does not introduce a released new dataset, codebase, or model artifact.

\item {\bf Crowdsourcing and research with human subjects}
    \item[] Question: For crowdsourcing experiments and research with human subjects, does the paper include the full text of instructions given to participants and screenshots, if applicable, as well as details about compensation (if any)? 
    \item[] Answer: \answerNA{}
    \item[] Justification: The work does not involve crowdsourcing or experiments with human participants.

\item {\bf Institutional review board (IRB) approvals or equivalent for research with human subjects}
    \item[] Question: Does the paper describe potential risks incurred by study participants, whether such risks were disclosed to the subjects, and whether Institutional Review Board (IRB) approvals (or an equivalent approval/review based on the requirements of your country or institution) were obtained?
    \item[] Answer: \answerNA{}
    \item[] Justification: The work does not involve crowdsourcing or research with human subjects.

\item {\bf Declaration of LLM usage}
    \item[] Question: Does the paper describe the usage of LLMs if it is an important, original, or non-standard component of the core methods in this research? Note that if the LLM is used only for writing, editing, or formatting purposes and does \emph{not} impact the core methodology, scientific rigor, or originality of the research, declaration is not required.
    \item[] Answer: \answerYes{}
    \item[] Justification: LLM usage is a core component of the method and is described in the abstract, Method section, and Experimental Setup.

\end{enumerate}